\documentclass[a4paper,11pt]{article}
\usepackage[utf8]{inputenc}
\usepackage[T1]{fontenc}

\usepackage{graphicx}
\usepackage[top=2.54cm,bottom=2.54cm, left=2.54cm, right=2.54cm]{geometry}
\usepackage{fancyhdr}
\usepackage{hyperref}
\usepackage{lastpage}
\usepackage{multicol}
\usepackage{titling}
\usepackage{cite}
\usepackage{amsmath}
\usepackage{amsfonts}
\usepackage{amssymb}
\usepackage{verbatim}
\usepackage{floatpag}
\usepackage{siunitx}
\usepackage{amsmath,lipsum}
\usepackage{multicol}
\usepackage{booktabs}
\usepackage{float}
\usepackage{pgfplots}
\usepackage{bm}
\usepackage{svg}
\usepackage{nameref}
\usepackage{longtable}
\usepackage{pdfpages}
\usepackage{subcaption}
\usepackage{tabularx}
\usepackage{multirow}
\usepackage{graphicx}
\usepackage{caption}
\usepackage{subcaption}
\usepackage{adjustbox}

\usepackage{graphicx}
\usepackage{caption}

\usepackage{lscape}
\usepackage{rotating}
\usepackage{tabularx}

\usepackage{parskip}
\usepackage{gensymb}
\usepackage{epsfig}
\usepackage{makecell}
\usepackage{enumitem}

\usepackage{mathtools}
\usepackage{amsmath}
\usepackage{algorithm}
\usepackage{algpseudocode}
\usepackage{subcaption}
\usepackage{placeins}

\newcommand{\varline}{\leavevmode\leaders\hrule height 3.2pt depth -2.8pt\hfill\kern0pt}
\newcommand{\rowterm}[1]{\makebox[8cm]{\varline \quad $\displaystyle #1$ \quad \varline}}
\usepackage{authblk}
\title{\textbf{\LARGE{An Introduction to Sparse Identification of Nonlinear Dynamics for Engineering Applications}}}

\author[1]{Yao Cheng Li$^*$}
\author[2,3]{Ana Larra\~{n}aga}
\author[2,3]{Steven L. Brunton}
\author[1]{Urban Fasel}

\affil[1]{\small Department of Aeronautics, Imperial College London, SW7 2AZ, United Kingdom}
\affil[2]{\small Department of Mechanical Engineering, University of Washington, Seattle, WA 98195, USA}
\affil[3]{\small NSF AI Institute in Dynamic Systems, University of Washington, Seattle, WA 98195, USA}
\affil[$*$]{{\footnotesize corresponding author: paul.li20@imperial.ac.uk}}

\begin{document}
\date{}
\maketitle
\vspace{-.5in}

\begin{abstract}
    Many engineering problems involve phenomena whose governing equations are poorly characterized or only partially known. Surrogate modeling techniques such as neural networks can capture the behavior of these systems, but they typically demand large training datasets that are difficult to obtain in engineering contexts and yield models with limited physical interpretability. The Sparse Identification of Nonlinear Dynamics (SINDy) method addresses both limitations by performing sparse regression over libraries of candidate nonlinear terms, recovering interpretable governing equations from comparatively small datasets. Although SINDy has been demonstrated extensively on canonical benchmark systems, its application to practical engineering problems is less widely documented. This tutorial introduces the SINDy method and progressively builds toward its main extensions, from noise-robust weak-form and ensembling-based variants to constrained and parametrizable formulations. The paper and the accompanying tutorial \footnote{Tutorial available at: \href{https://github.com/paullililili/SINDy4Engineers}{https://github.com/paullililili/SINDy4Engineers}} is organized in three parts: the first introduces the standard SINDy algorithm and progressively extends it, inviting readers without prior knowledge to follow each step and adapt the methods to their own problems; the remaining two parts present detailed case studies on (1) the system identification of an unmanned aerial vehicle and (2) a chaotic thermosyphon heat exchanger. Through these examples, we aim to demonstrate that SINDy is simple to implement yet flexible enough to serve as a valuable identification tool for advanced engineering applications.
\end{abstract}

\section{Introduction}
Consider the task of modeling the reaction dynamics of a continuous stirred tank reactor using live data, in order to develop a controller that adjusts the heat transfer rate into the tank to control temperature and reactant concentration \cite{bhadriraju_machine_2019}. Given the strong nonlinearities involved, one could approach this challenge with a neural network, but such models capture input--output correlations rather than the governing equations themselves, offering little insight into the underlying mechanisms~\cite{Rudin2019}. An alternative is to recover the governing dynamics in interpretable form, so that the resulting model itself informs the design decision. This is the motivation behind the Sparse Identification of Nonlinear Dynamics (SINDy) framework~\cite{brunton_sindy_2016}. High-stakes engineering decisions require an understanding of a system's governing physics, not merely a prediction of its behavior. SINDy addresses this need through sparse regression on time-series data, identifying the few nonlinear terms that best characterize the system's dynamics. Despite its simplicity, the framework has proven effective across a wide range of applications, such as fluid dynamics~\cite{Fukami_Murata_Zhang_Fukagata_2021}, neuroscience~\cite{Delabays2025}, and epidemiology~\cite{Horrocks2020}.

Researchers have repeatedly shown that SINDy can identify the dynamics of both ordinary differential equations (ODEs)~\cite{brunton_sindy_2016} and partial differential equations (PDEs)~\cite{rudy_PDE-FIND_2017}. Common examples include the Lorenz system and the Van der Pol oscillator for ODEs, and the inviscid Burgers equation or a 3D reaction-diffusion model for PDEs. These benchmarks, however, rarely match the conditions found in practice. Measured engineering data are hardly ever clean: sensors introduce noise, sampling is often sparse or irregular, data can come from different sources and vary in quality, acquisition is limited by cost or operational constraints, and the systems of interest are frequently high-dimensional. These limitations are rarely treated from an engineering perspective in the literature, yet, each has an applicable SINDy extension: the weak or integral formulation for noisy and sparsely sampled data~\cite{messenger_WSINDy-ODE_2021, messenger_WSINDy-PDE_2021,schaeffer_Weak-SINDy_2017, reinbold_Weak-SINDy_2020}, active or ensemble SINDy for limited or noisy measurements~\cite{larranaga2026,fasel_eSINDy_2022}, multi-fidelity SINDy for heterogeneous data~\cite{zacchei2026multifidelity}, constrained scenarios~\cite{Loiseau2017jfm, champion_SR3_2020, kaptanoglu2021promoting}, and its combination with autoencoders for high-dimensional systems~\cite{champion_Autoencoder-SINDy_2019,bakarji_Time-Delay-Autoencoder-SINDy_2023, gao_Bayesian-Autoencoder-SINDy_2022, conti_VINDy_2024}.

\begin{figure}[t]
    \centering
    \includegraphics[width=1\linewidth]{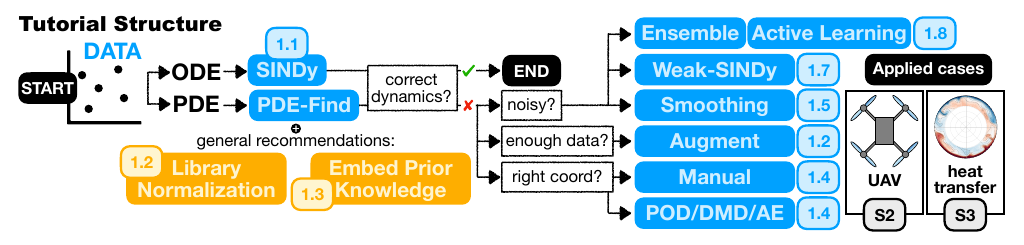}
    \vspace{-0.8cm}
    \caption{\small \textbf{SINDy Tutorial Structure}. The repository (\href{https://github.com/paullililili/SINDy4Engineers}{hosted in GitHub}) is organized into three distinct sections. Section 1 provides a self-contained implementation of all SINDy methods without class dependencies. Sections 2 and 3 present applied examples that use the official SINDy package, PySINDy, to learn the dynamics of an unmanned aerial vehicle and a chaotic flow, respectively. Numbers in the figure indicate the corresponding tutorial notebook in which each component is explained.}
    \vspace{-0.0cm}
    \label{fig:pipeline}
\end{figure}

This tutorial builds an understanding of SINDy and its main extensions from the ground up (see Fig~\ref{fig:pipeline}), separating the method itself from the software choices used to implement it \cite{silva_pysindy_2020, kaptanoglu2021pysindy}. The accompanying notebooks present the mathematics alongside the corresponding code, keeping the two distinct before applying it to several ODE and PDE examples. It then applies these techniques to two engineering problems: (1) an unmanned aerial vehicle and (2) a chaotic thermosyphon flow.

\section[SINDy]{Sparse Identification of Nonlinear Dynamics}
\label{sec: SINDy Introduction}
SINDy is a data-driven method for identifying the governing equations of a system directly from time-series measurements of its state. It applies sparse regression to select a small set of terms from a library of candidate functions, identifying a set of differential equations that describe how the system evolves~\cite{brunton_sindy_2016}. Because the resulting model is an explicit equation rather than a black-box model, it is physically interpretable and tends to extrapolate better to unseen conditions than over-parametrized models such as neural networks.

Building from this motivation, suppose that we have a dynamical system defined by a state vector with $n$ variables, $\boldsymbol{x}(t) \in \mathbb{R}^n$, such that
\begin{equation}
    \frac{\mathrm{d}}{\mathrm{d}t} \boldsymbol{x}(t) = \boldsymbol{f}(\boldsymbol{x}(t)),
\end{equation}
where \textbf{$f$} is potentially a nonlinear function. In an RC circuit (resistor-capacitor) example, the state can be the voltage decay whose dynamics are governed by
\begin{equation}
    \frac{\mathrm{d}V}{\mathrm{d}t} = f(V(t)) = -\frac{V(t)}{RC}.
\end{equation}
SINDy assumes that $\boldsymbol{f}$ can be approximated as a linear combination of nonlinear candidate functions $\theta$ collected in a library matrix $\Theta(X) \in \mathbb{R}^{m \times l}$, where $X \in \mathbb{R}^{m \times n}$ is the collated time-series data matrix over $m$ measurements to be used in $l$ number of candidate functions (see \textbf{Tutorial 1.1}). The library is chosen from prior knowledge of the system, typically starting with simpler terms (such as polynomials of the state variables) and adding complexity as needed (for instance, trigonometric terms). The objective is to find a sparse coefficient matrix $\Xi \in \mathbb{R}^{l \times n}$ such that
\begin{equation}
    \dot{X} \approx \Theta(X)\,\Xi.
\end{equation}
Higher-degree polynomial terms can make $\Theta(X)$ ill-conditioned, meaning that small perturbations in the data (e.g., measurement noise) can produce large changes in the computed coefficients. This is quantified by the condition number $\kappa(\Theta)$: a large value indicates strong amplification of such errors, which in turn increases sensitivity to noise and to the choice of the sparsification threshold (see \textbf{Tutorial 1.2}). It is also good practice to normalize the columns of $\Theta(X)$ when the candidate functions have very different magnitudes, as this affects the sparse regression step to solve $\dot{X} = \Theta(X)\Xi$, which is the core idea of SINDy.

Sparsity is the key assumption: most physical systems are governed by only a few terms, so we seek a solution in which most coefficients are exactly zero. Originally, Brunton et al.~\cite{brunton_sindy_2016} proposed sequentially thresholded least squares (STLSQ), a simple approach that first solves an ordinary least squares regression and then zeros out any coefficient falling below a chosen threshold $\lambda$ (interpreting small coefficients as spurious). The ordinary least squares regression and thresholding steps are then repeated on the remaining terms until a sparse solution is found. The problem is solved for each state variable $x_i$, identifying the sparse coefficient vector $\xi_i$ (the $i$-th column of $\Xi$) by minimizing
\begin{equation}
\min_{\xi_i} \| \dot{X}_i - \Theta(X)\,\xi_i \|^2,
\end{equation}
where $\dot{X}_i$ is the corresponding column of $\dot{X}$. In practice, the sparsification step is applied after the least squares regression by setting all coefficients with a magnitude below $\lambda$ to zero. If $\lambda$ is too high, we threshold out the key dynamics, resulting in under-fitting. If $\lambda$ is too low, we include excess terms that overfit to the dataset. Hence, a right balance must be found.

To correctly identify the governing equations, SINDy requires data that sufficiently captures the behavior of the system. Trajectories that are too short, sampled too infrequently, or initialized from a limited range of initial conditions may fail to resolve the full dynamics and lead to an incorrect identification. 
When data is limited, incorporating prior physical knowledge can help guide the sparse regression toward the correct equations, which can be done for example by constraining the candidate function library~\cite{Loiseau2017jfm, champion_SR3_2020, Zheng2019ieeeacess}, or by selecting a model based on coefficient stability \cite{maddu_stability-selection_2022} (see \textbf{Tutorial 1.3}).

\subsection{Coordinate Basis Selection}
In many practical applications, the observed data collected from experimentation or numerical simulations may not directly measure the state(s) of the system. For instance, if we track an undamped pendulum using video cameras, the motion-tracking software may return the $xy$ position of the mass rather than its angular displacement. Learning the system dynamics using $xy$ coordinates would require learning the rotation transformation simultaneously, whereas directly learning from the angular states would yield the critical and yet sparse governing equation of $\ddot{\theta} = -g\sin(\theta)$ (see \textbf{Tutorial 1.4}). If a system's intrinsic coordinate basis is known, coordinate transformation should always be performed as part of data pre-processing.

Whilst SINDy excels at discovering dynamics in low-dimensional problems, its candidate function library scales combinatorially with the number of states, making it computationally intractable for high-dimensional systems. In such cases, SINDy can be combined with dimensionality-reduction methods such as Proper Orthogonal Decomposition (POD) or Dynamic Mode Decomposition (DMD), allowing it to discover dynamics in a lower-dimensional latent space. 
For instance, the flow field around a cylinder can be decomposed into several wake shedding modes using POD~\cite{brunton_sindy_2016}, after which SINDy can identify the governing equations of the reduced system.

In POD, measurements at each snapshot in time $\boldsymbol{x}(t_j)\in \mathbb{R}^{m}$ are stacked as columns of a data matrix $\boldsymbol{X} \in \mathbb{R}^{m \times n}$, where each column is one of $n$ snapshots of the $m$-dimensional state such that $\boldsymbol{X} = [\boldsymbol{x}(t_1) \; \boldsymbol{x}(t_2) \dots \; \boldsymbol{x}(t_n)]$. It extracts a set of orthogonal spatial modes from this matrix using the singular value decomposition (SVD)~\cite{weiss2019tutorial}
\begin{equation}
    \boldsymbol{X} = \boldsymbol{U} \boldsymbol{\Sigma} \boldsymbol{V}^T,
\end{equation}

where the columns of the orthonormal matrix $\boldsymbol{U} \in \mathbb{R}^{m \times m}$ are the spatial modes, the diagonal matrix $\boldsymbol{\Sigma} \in \mathbb{R}^{m \times n}$ holds the singular values $\sigma_i$ that rank each mode by its amplitude, and the orthonormal rows of $\boldsymbol{V}^T \in \mathbb{R}^{n \times n}$ describe the temporal evolution of the modes~\footnote{In this tutorial, we assume that typical engineering data is real-valued. A more general SVD formulation uses the Hermitian transpose $\boldsymbol{V}^H$ for complex-valued data, or adjoint $\boldsymbol{V}^*$ for non-standard inner products.}. The POD modes are arranged in descending amplitude, from most to least energetic, in the column space of $\boldsymbol{U}$. The first $r$ columns $\boldsymbol{U}_r$ can then be used to project the data onto a lower-dimensional space via $\boldsymbol{Z} = \boldsymbol{U}_r^{T}\boldsymbol{X}$. Here, $\boldsymbol{Z} \in \mathbb{R}^{r \times n}$ contains the temporal evolution of the $r$ POD modes, which can then be used with SINDy to discover the dynamics in the latent space (see \textbf{Tutorial 1.4}). Unlike POD, DMD computes a linear mapping matrix $\boldsymbol{A}$ that relates the snapshot matrix $\boldsymbol{X} = [\boldsymbol{x}(t_1) \; \boldsymbol{x}(t_2) \dots \; \boldsymbol{x}(t_{n-1})]$ to its time-shifted counterpart $\boldsymbol{Y} = [\boldsymbol{x}(t_2) \; \boldsymbol{x}(t_3) \dots \; \boldsymbol{x}(t_n)]$ via $\boldsymbol{Y=AX}$. Using reduced-rank regression to compute $\boldsymbol{A}$ \cite{loiseau_data-driven_2020}, the eigenvectors $\boldsymbol{\Psi}\in \mathbb{R}^{m \times r}$ of $\boldsymbol{A}$ can be used to project the data to an $r$-dimensional space $\boldsymbol{Z} = \boldsymbol{\Psi}^{\dagger} \boldsymbol{X}$, which can again be used with SINDy to discover dynamics in a low-dimensional latent space.

POD and DMD are both linear dimensionality reduction methods that often require many modes to capture the dominant physics. Autoencoders can be considered a nonlinear extension of these methods, compressing the same amount of variance (or energy in the fluid flow example) into significantly fewer latent variables~\cite{lee2020model}. Specifically, an autoencoder is a type of neural network that learns a nonlinear mapping through an encoder $\varphi: \mathbb{R}^m \to \mathbb{R}^r$ and a decoder $\psi: \mathbb{R}^r \to \mathbb{R}^m$ where $\boldsymbol{z}(t_j) = \varphi(\boldsymbol{x}(t_j))$ and $\boldsymbol{x}(t_j)=\psi(\boldsymbol{z}(t_j))$, enabling it to capture more complex structures with fewer latent variables than linear methods. Combined with SINDy, as first demonstrated by Champion et al.~\cite{champion_Autoencoder-SINDy_2019}, this approach enables the simultaneous learning of the dimensionality reduction mapping and the sparse governing dynamics within the latent space (see \textbf{Tutorial 1.4}).

\subsection{Noise Robustness}
Obtaining accurate time derivatives is one of the main challenges when using SINDy, as they are rarely measured directly and must instead be computed numerically using techniques such as finite differences (see \textbf{Tutorial 1.1}). When the data is noisy, however, direct numerical differentiation becomes unreliable, as illustrated by the following example. Let us define
\begin{equation}
y = f(x) + \epsilon(\sigma)
\end{equation}
where $\epsilon$ is an additive Gaussian noise component with a standard deviation of $\sigma$. Specifically, let us define $\epsilon$ as a single sine wave with frequency $\omega_{\text{noise}}$ and amplitude $A$ such that
\begin{subequations}
\begin{align}
    y &= f(x) + A \sin(\omega_{\text{noise}} x), \\
    \frac{d^n y}{dx^n} &= f^{(n)}(x) +
    \begin{cases}
        (-1)^{\frac{n}{2}} A \omega_{\text{noise}}^n \sin(\omega_{\text{noise}} x) & \text{for } n \text{ even} \\
        (-1)^{\frac{n-1}{2}} A \omega_{\text{noise}}^n \cos(\omega_{\text{noise}} x) & \text{for } n \text{ odd}
    \end{cases}
\end{align}
\end{subequations}
Therefore, the higher the frequency, the larger the amplitude of the noise component in the derivative, which is further amplified by using higher-order derivatives. Some methods can mitigate this, such as data smoothing algorithms and differentiation with total variation regularization \cite{chartrand_numerical_2011} (see \textbf{Tutorial 1.5}), as well as weak-form SINDy formulations \cite{messenger_WSINDy-ODE_2021, messenger_WSINDy-PDE_2021} that bypass point-wise differentiation altogether (see \textbf{Tutorials 1.7.1 and 1.7.2}).

\subsection{Controller Dynamics Identification}
A natural extension of SINDy is SINDy with Controls (SINDyc)~\cite{brunton_SINDyc_2016}, which enables the discovery of parameterized systems or systems subjected to forcing or control inputs $\boldsymbol{u}(t) \in \mathbb{R}^{q}$ such that
\begin{equation}
    \frac{d}{dt} \boldsymbol{x}(t) = \boldsymbol{f}(\boldsymbol{x}(t), \boldsymbol{u}(t)).
\end{equation}
In SINDyc, the candidate function library $\Theta(X,U)$ is extended to include controls $U$. For a polynomial based library, the candidate functions are $x,\ y,\ z,\ u,\ x^2,\ xy,\ \dots,\ zu^2,\ u^3$. Once constructed, the exact same sparse regression problem is repeated to solve
\begin{equation}
    \dot{X} \approx \Theta(X,U) \Xi.
\end{equation}
The learned model can for example be used with a receding horizon model predictive controller (MPC), which is an optimal control strategy that uses a model of the system to predict the evolution of the state over a finite horizon~\cite{kaiser_SINDy-MPC_2018} (see \textbf{Tutorial 1.6}).

Alternatively, SINDy can be combined with reinforcement learning (RL), a framework in which an agent learns a control policy through trial-and-error interaction with its environment by maximizing a cumulative reward. Standard deep RL is notoriously data-hungry and yields black-box policies that are not interpretable. SINDy-RL instead learns sparse, interpretable surrogate models of the system dynamics, the reward function, and the control policy~\cite{zolman_SINDy-RL_2025}. By replacing expensive environment interactions with cheap evaluations of the learned dynamics model, it improves sample efficiency whilst obtaining compact, interpretable policies.

\subsection{Ensembling \& Active Learning}
Finally, uncertainty quantification is a desirable capability, especially when operating in the low data regime. Ensembling methods can tackle this, combining multiple models trained on different subsets of the data, most commonly through bootstrap aggregation (bagging). In the SINDy setting, this is known as Ensemble-SINDy (E-SINDy)~\cite{fasel_eSINDy_2022}, which bootstraps from the data samples and, optionally, from the candidate library terms. Each of the $m$ ensemble members is then fitted independently,
\begin{equation}
    \dot{X}^{(k)} \approx \Theta\big(X^{(k)}\big)\,\Xi^{(k)}, \qquad k = 1,\dots,m,
\end{equation}
and the resulting coefficients are aggregated by their mean to give a bagged estimate, while the median yields a more outlier-robust variant (bragging),
\begin{equation}
    \bar{\Xi} = \frac{1}{m}\sum_{k=1}^{m}\Xi^{(k)} \quad\text{(bagging)}, \qquad \tilde{\Xi} = \operatorname{median}_{k}\,\Xi^{(k)} \quad\text{(bragging)}.
\end{equation}
The model ensemble can provide an uncertainty estimate for both the predictions and the coefficient identification, making it possible to track which coefficients appear most consistently across the ensemble (see \textbf{Tutorial 1.8.1}). In fact, such uncertainty estimates can be used as a sampling criterion in active learning scenarios, where gathering new data is expensive and each sampled time-series must be used efficiently~\cite{larranaga2026}. Rather than sampling long periods of time, short trajectories are distributed across different regions of the state space, enabling new dynamics to be discovered at a lower data cost (see \textbf{Tutorial 1.8.2}).
\section[UAV System Identification]{ODE Case Study: UAV System Identification}

In recent years, unmanned aerial systems have seen increased deployment across a wide range of applications, from cargo and personnel transportation to monitoring, surveillance, mapping, and inspection, establishing themselves as a cornerstone enabling technology~\cite{annaswamy_control_2024}. This calls for rapid innovation and iterations to seek out unique design solutions that address various requirements. For instance, a hybrid fixed-wing vertical takeoff and landing (VTOL)-capable unmanned aerial vehicle (UAV) exploits both the long range and endurance of a fixed wing, while offering the flexible operational capabilities of a rotorcraft vehicle~\cite{olsson_Wintra-SysID_2021}. However, the dynamics of such vehicles are challenging to model due to effects such as wing-propeller interactions or vertical to forward flight transitions. Furthermore, UAVs have become widely accessible and can be rapidly prototyped and tested, making flight test data increasingly available compared to high-fidelity computational simulations or wind tunnel experiments.  This accessibility motivates the use of data-driven methods to identify UAV dynamics purely from time-series measurements, or to correct existing low to medium-fidelity models using system identification methods. 

The common approach to UAV system identification involves parametric estimation using a known analytical model~\cite{baskin_frequency-domain_2022, pounds_system_2007, simmons_nonlinear_2023}. Recent developments in deep learning have also spurred the application of black-box neural networks in the identification process~\cite{harris_aircraft_2016, pairan_neural_2020} to circumvent the need for known models. In our tutorial, we present a constrained SINDy approach that enforces known equation structures and coefficients, whilst still allowing SINDy to learn additional terms that describe physics without known priors (see \textbf{Tutorial 2}). The use of constrained regression, noisy data smoothing, and custom function library design methods introduced in the previous tutorials will be applied here.

\begin{figure}[tbh]
	\centering
	\includegraphics[width=1.0\textwidth]{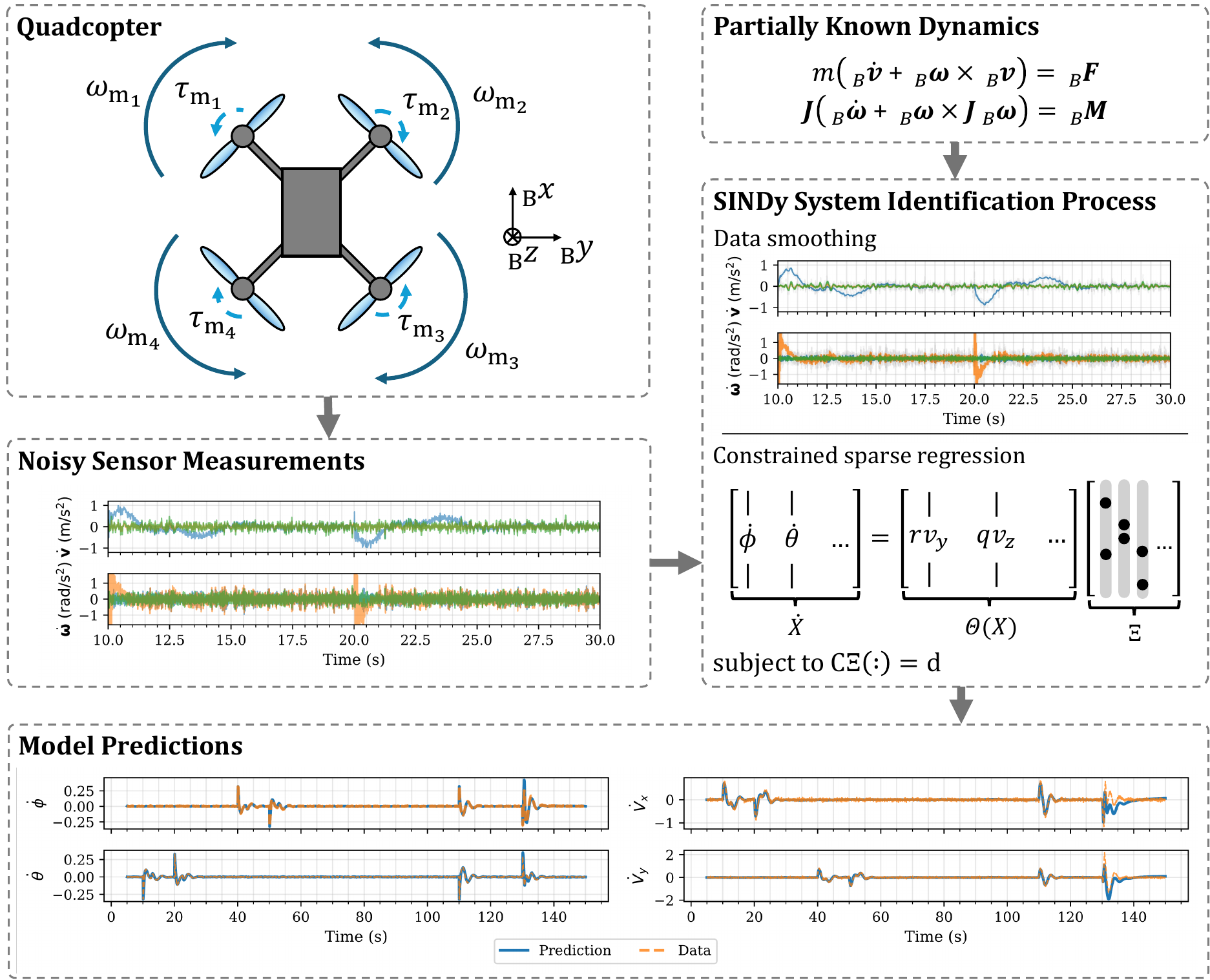}
    \vspace{-0.6cm}
	\caption{\small Schematic of the UAV system identification process. We first extract sensor measurements from simulated flight test data before applying smoothing techniques. A constrained sparse regression approach is then taken using partial knowledge of the model to identify the governing equations and predict future trajectories with it.}
	\label{fig: Quadrotor top-down view}
\end{figure}

\subsection{Quadcopter Model and Simulation}

In this tutorial, we use a quadcopter UAV configuration, as shown in Figure \ref{fig: Quadrotor top-down view}, propelled by four motors and propellers in a cross layout. All four motors contribute equally to the vertical thrust of the UAV, and differential thrust contributes to the rolling moments $M_x$ or pitching moments $M_y$. Opposing pairs of motors rotate in the same direction to generate equal motor torque $\tau_m$, and differential control pairs contribute to yawing moments $M_z$. The full six degrees of freedom system defining a quadcopter's kinematics, defined by its translational velocities $\prescript{}{B}{\boldsymbol{v}} = \begin{bmatrix} v_x & v_y & v_z \end{bmatrix}^T$ and angular velocities $\prescript{}{B}{\boldsymbol{\omega}} = \begin{bmatrix} p & q & r \end{bmatrix}^T$. Whilst most applied forces and moments, such as those generated by thrust inputs, are oriented in the body frame of reference (denoted by $\prescript{}{B}{(\cdot)}$), gravitational force is oriented in the inertial frame of reference. As a result, the governing equation contains rotational matrices and cross product terms from rigid body kinematics shown in Figure~\ref{fig: Quadrotor top-down view}. The full expanded dynamics of the UAV system is not easily discoverable with the standard SINDy implementation and are defined as
\begin{subequations}
    \label{eqn: UAV expanded dynamics}
    \begin{align}
        \dot{\phi} &= p + \tan(\theta) \left[ q \sin(\theta) + r \cos(\phi) \right], \\
        \dot{\theta} &= q \cos(\phi) - r \sin(\phi), \\
        \dot{\psi} &= \sec(\theta) \left[ q \sin(\theta) + r \cos(\phi) \right], \\
        {\dot{v}_x} &= (r v_y - q v_z) - g \sin(\theta) + F_x / m, \\
        \dot{v}_y &= (p v_z - r v_x) + g \cos(\theta) \sin(\phi) + F_y / m, \\
        \dot{v}_z &= (q v_x - p v_y) + g \cos(\phi) \cos(\theta) + F_z / m, \\
        \dot{p} &= \left[ M_x + qr(I_{yy} - I_{zz}) \right] / I_{xx}, \\
        \dot{q} &= \left[ M_y + pr(I_{zz} - I_{xx}) \right] / I_{yy}, \\
        \dot{r} &= \left[ M_z + pq(I_{xx} - I_{yy}) \right] / I_{zz}.
    \end{align}
\end{subequations}

We can omit the position states of the UAV for this tutorial, since they do not affect the dynamics when ground effect is not present and wind is negligible. Whilst complex, these functions are known from rigid body kinematics, and only the applied forces $F$ and moments $M$ remain unknown. We can therefore use a constrained regression approach with constrained SR3 algorithm~\cite{champion_SR3_2020} to allow SINDy to discover the unknown forces and moments, while retaining known components of the dynamics. In the tutorial, we explore two separate approaches: the construction of a tailored candidate function library to include all non-externally applied force and moment functions in Equations \ref{eqn: UAV expanded dynamics}, and second, applying linear equality constraints on known coefficients in Equations \ref{eqn: UAV expanded dynamics}.

We generate the training and validation data using Mathwork's Parrot Minidrone simulation in Matlab 2025b, which is a medium-fidelity simulation that models certain physics such as blade flapping, pitch and roll rotor damping, and body drag, but ignores higher-fidelity physics such as unsteady aerodynamics with time-scales too small to be significant here~\cite{pounds_modelling_2006, riether_agile_2016,pounds_system_2007}.  Furthermore, it simulates onboard Kalman filter-based state estimators, which use sensor data from a combination of simulated IMU sensors and a mimicked visual odometry sensor for position and velocity drift corrections. Furthermore, an onboard cascaded-PID controller controls the motor RPM to match reference position states and yaw angle. To obtain data that is sufficiently rich in dynamics that SINDy may learn from, we generate a series of decoupled and coupled step and chirp reference signals that the controller must follow. Finally, we collate all time-series state measurements into a data matrix $X \in \mathbb{R}^{m \times n}$ and control matrix $U \in \mathbb{R}^{m \times q}$.

\subsection{Constrained SINDy Learning}

Since many of the known functions, specifically ones that originate from the application of the rotation matrices, within Equations \ref{eqn: UAV expanded dynamics} cannot be built from simple basis function libraries such as polynomial or Fourier libraries, we define custom functions using the appropriate terms such that
\begin{equation}
    \Theta_\text{custom}(X) = \begin{bmatrix}
        \rowterm{p + \tan(\theta) \left( q \sin(\phi) + r \cos(\phi) \right)} \\
        \rowterm{q \cos(\phi) - r \sin(\phi)} \\
        \rowterm{\sec(\theta) \left( q \sin(\phi) + r \cos(\phi) \right)} \\
        \rowterm{-\sin(\theta)} \\
        \rowterm{\cos(\theta) \sin(\phi)} \\
        \rowterm{\cos(\phi) \cos(\theta)} \\
        \vdots
    \end{bmatrix}^T.
\end{equation}
In addition to the known terms present in the Equations \ref{eqn: UAV expanded dynamics}, we include additional linear velocity squared terms, and pitch and roll rates to allow SINDy to learn the effects of body drag and rotor damping, respectively. We can further introduce a 3rd order polynomial basis library for control inputs only such that $\Theta_\text{polynomial}(U) = \begin{bmatrix} \boldsymbol{u}_1(t) & \boldsymbol{u}_2(t) & \boldsymbol{u}_3(t) & \boldsymbol{u}_4(t) & \dots & (\boldsymbol{u}_4(t))^3 \end{bmatrix}$ where $\boldsymbol{u}_i(t)$ corresponds to the time-series input for the $i$th motor. This library allows SINDy to discover the control terms present in the system. Finally, we concatenate the libraries together such that
\begin{equation}
    \Theta(X, U) = \begin{bmatrix} \Theta_\text{custom} (X) & \Theta_\text{polynomial} (U) \end{bmatrix},
\end{equation}
where the library matrix $\Theta(X,U) \in \mathbb{R}^{m \times l}$ contains $l$ candidate functions. Next, to enforce sparsity or enforce known coefficients as shown in Section \ref{sec: SINDy Introduction}, we can constrain known functions and equation structure by creating linear constraints $C{\Xi(:)}=d$. In the tutorial, we programmatically create the constraints. For instance, to enforce unity on $p + \tan(\theta) \left( q \sin(\phi) + r \cos(\phi) \right)$ for $\dot{\phi}$, we set our first constraint as
\begin{equation}
    \label{eqn: known coefficient constraint}
    C_{1,j} = \begin{cases} 
        1 & \text{if } j = 1, \\
        0 & \text{if } j \neq 1,
    \end{cases}
\end{equation}
with $d_1 = 1$. We must also enforce sparsity on this function for all other states such that
\begin{align}
    C_{i,j} &= \begin{cases} 
        1 & \text{if } j = (i-1)l+1, \\
        0 & \text{if } j \neq (i-1)l+1,
    \end{cases}
\end{align}
and $d_i = 0$ for $i \in [2,n]$. The same approach can be applied to other custom functions, such as functions with gravitational acceleration constant $g$, to enforce both coefficient and sparsity. In some cases, however, we omit the application of Equation \ref{eqn: known coefficient constraint} to allow SINDy to learn the coefficients. For instance, for the cross-coupling terms of angular rates $qr$, $pr$, and $pq$, we allow SINDy to learn the coefficients which are a function of the UAV's moments of inertia $I_{xx}$, $I_{yy}$, and $I_{zz}$. Further constraints can also be enforced, such as control input symmetry. For instance, the forward and rear sets of motors should have equal and opposite pitching moments.

In this tutorial, the combination of defining custom candidate functions and constraining known coefficients and sparsity yields an interpretable model that matches the expected equation structure and produces accurate predictions despite the noisy data, as shown in Figure \ref{fig: Quadrotor top-down view}. While demonstrated here on a UAV dataset, the approach generalizes naturally to other system identification problems with partially known coefficients and equation structure, and we encourage the reader to test it on their own engineering datasets. For readers interested specifically in UAVs, there are also other recent approaches to applying SINDy in UAV system identification. For example, Osman et al.~\cite{osman_adaptive_2025} who presented an application of an adaptive SINDy framework integrated with a Lyapunov-based Model Predictive Controller to produce a real-time controller for a VTOL-capable flying wing. Manaa et al.~\cite{manaa_data-driven_2024} have also presented an application of SINDy to learn the governing dynamics of a quadrotor system constrained to three degrees of freedom only, using a combination of polynomial and Fourier libraries.
\section[Chaotic Thermosyphon Flow]{PDE Case Study: Chaotic Thermosyphon Flow}

\begin{figure}[t]
	\centering
	\includegraphics[width=\textwidth]{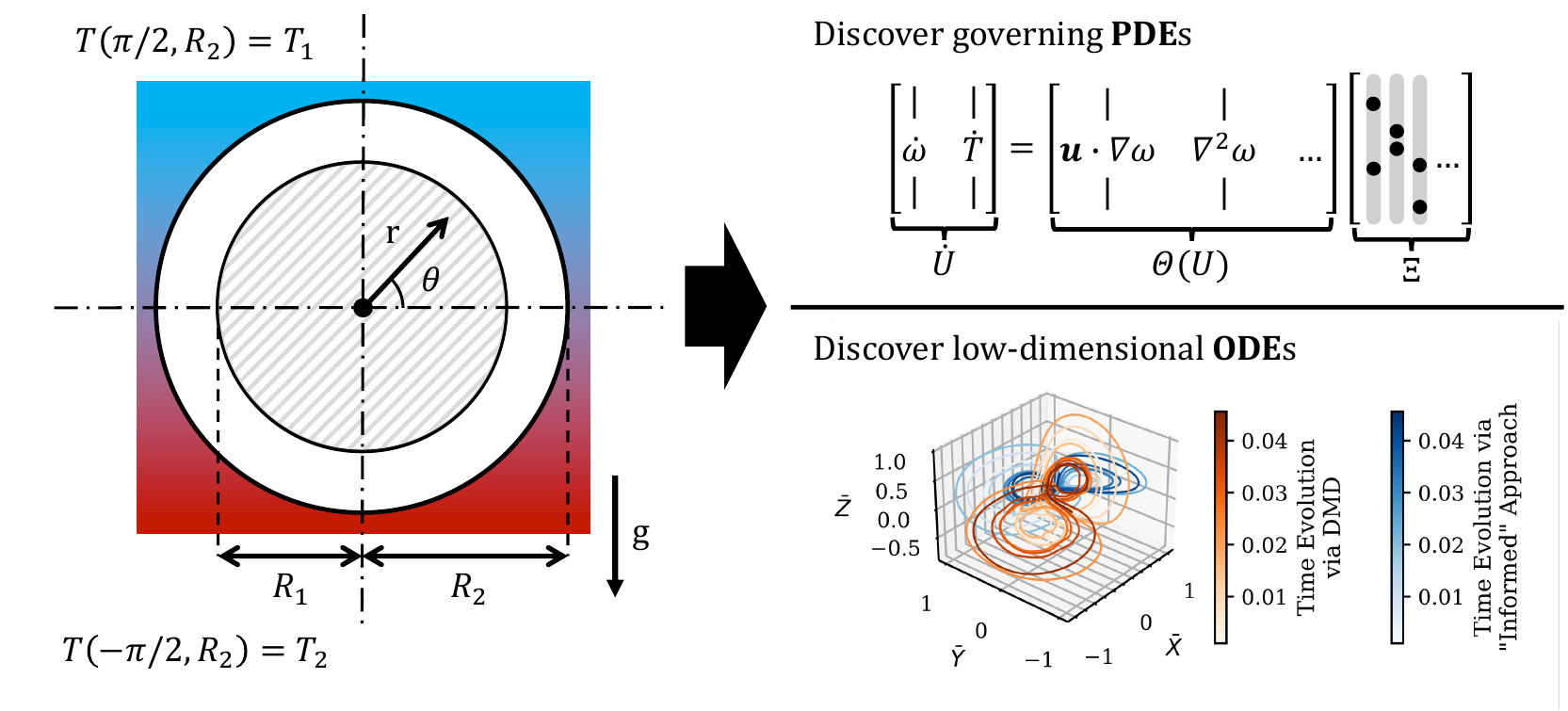}
    \vspace{-0.8cm}
	\caption{\small A thermosyphon contains convection-driven flow within its concentric inner and outer walls, subjected to a Dirichlet temperature boundary condition on its outer wall. Its inner wall is adiabatic. The governing PDEs may be discovered using PDE-FIND with a differential operator library. We may also discover a lower-dimensional latent space attractor (shown in normalized coordinates) using a fully data-driven approach that combines SINDy with DMD or using an "informed" approach where we know the variables to measure or reduce to.}
	\label{fig: Thermosyphon geometry}
\end{figure}

In this section of the tutorial, we apply SINDy to identify the governing PDEs of a flow within a thermosyphon shown in Figure~\ref{fig: Thermosyphon geometry}, and learn a lower-dimensional representation that can be used to analyze and predict the flow's evolution. PDEs present unique challenges to the identification process. For example, directly adding partial derivatives into the candidate function library like in PDE-FIND \cite{rudy_PDE-FIND_2017} results in high degrees of collinearity between functions. Furthermore, SINDy is primarily designed to discover equations where the time derivatives of its states can be explicitly defined, and its terms are spatially invariant. If such criteria are not met, such as in this tutorial (see Equations \ref{eqn: thermosyphon equations}), we have to develop alternative workarounds. Additionally, we are often interested in identifying a nonlinear dynamical system that lies within a lower-dimension latent space from the spatiotemporal solution, which is useful to analyze and predict future evolution of the solution fields.

A thermosyphon is a passive heat exchanger that relies on convection-driven flow to transport heat around the geometry. Early studies found that self-excited oscillations can occur in closed-loop thermosyphon geometries due to the interplay between buoyant and viscous forces~\cite{keller_periodic_1966, creveling_stability_1975}. This was later shown to have similar bifurcation behaviors as the chaotic Lorenz system~\cite{yorke_lorenz-like_1987}. More recently, Huang et al.~\cite{huang_convective_2023} derived a low-dimensional model from Naiver-Stokes Boussinesq equations, which treats the flow as incompressible but assumes that its temperature $T$ varies linearly with its density, using a truncated Fourier-Laurent expansion. Loiseau~\cite{loiseau_data-driven_2020} instead implemented a rank-constrained DMD to reduce the solution's dimensionality before applying constrained SINDy to identify a Lorenz like system. The low-dimensional time-series trajectories from both approaches are shown in Figure~\ref{fig: Thermosyphon geometry}. In this tutorial, we follow the flow setup in~\cite{huang_convective_2023} to demonstrate that the derived analytical system can be discovered purely from data using SINDy, as well as demonstrating an alternative approach to the fully data-driven DMD method presented in~\cite{loiseau_data-driven_2020}.

 The annulus geometry of the thermosyphon is presented in Figure \ref{fig: Thermosyphon geometry}. The heat exchanger is subjected to a temperature Dirichlet boundary condition on its outer walls that varies linearly with the vertical height, and the inner wall is adiabatic. The Navier-Stokes Boussinesq equations can be reformulated into the divergence-free vorticity-streamfunction equations 
  \vspace{1pt}
\begin{subequations}
    \label{eqn: thermosyphon equations}
    \begin{align}
        \frac{\partial \omega}{\partial t} + \boldsymbol{u} \cdot \nabla \omega &= \mathrm{Pr} \nabla^2 \omega + \mathrm{Pr} \, \mathrm{Ra} \left( \frac{\partial T}{\partial r} \cos \theta - \frac{1}{r} \frac{\partial T}{\partial \theta} \sin \theta \right), \\
        \frac{\partial T}{\partial t} + \boldsymbol{u} \cdot \nabla T &= \nabla^2 T, \\
        \label{eqn: theromsyphon streamfunction}
        -\nabla^2 \psi &= \omega,
    \end{align}
\end{subequations}
which are parameterized with the Rayleigh number $\mathrm{Ra}$, which defines the ratio between buoyant and viscous forces, and Prandtl number $\mathrm{Pr}$, which defines the ratio between momentum and viscous diffusivity. At low Rayleigh numbers, any transients in the flow quickly decay away and the flow returns to a zero-circulation static state. Past a critical Rayleigh number, the flow undergoes a pitchfork bifurcation and settles into either a clockwise or counter-clockwise constant circulation. At even higher Rayleigh numbers, the flow transitions into a chaotic regime and exhibits aperiodic flow reversal behavior. The ability to predict such bifurcation points is critical to any engineer designing such a heat exchanger (see \textbf{Tutorial 3.1}).

\begin{figure}[t]
	\centering
	\includegraphics[width=1.0\textwidth]{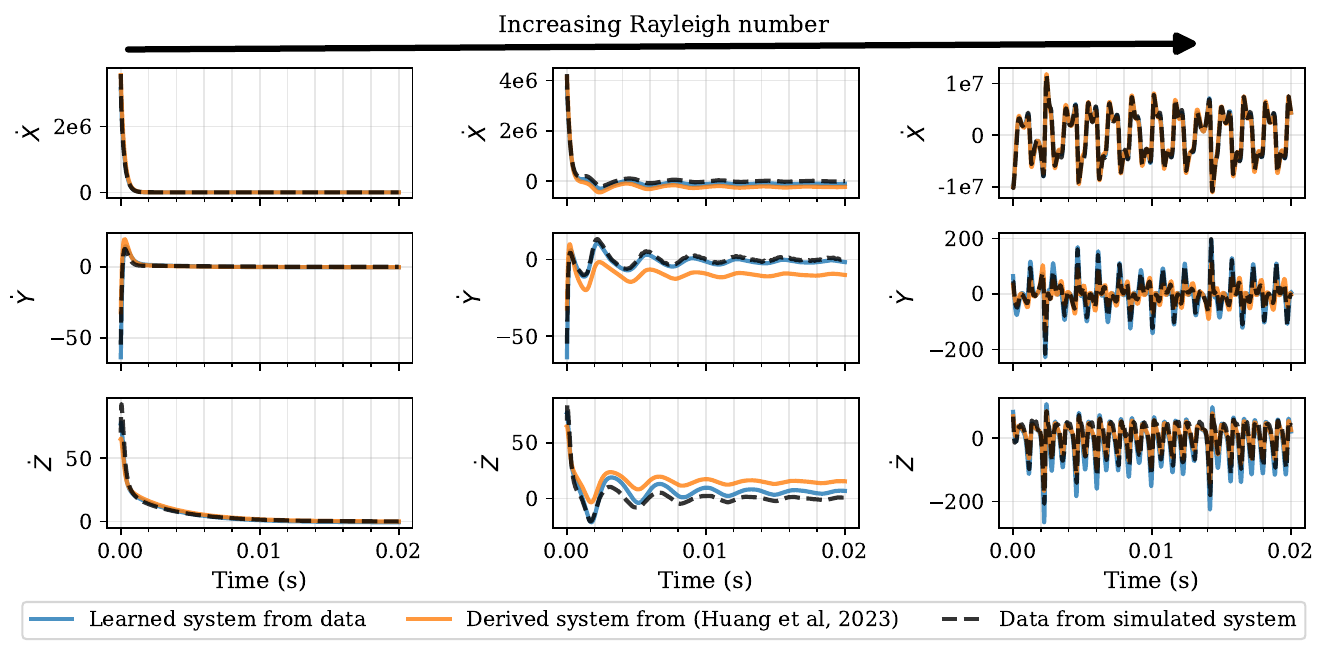}
    \vspace{-0.8cm}
	\caption{\small A comparison of the state derivative predictions between the learned SINDy model and derived model against simulated data for increasing Rayleigh number.}
	\label{fig: thermospyphon ode comparison}
\end{figure}

In an "informed" approach, we can extract low-dimensional states from the spatiotemporal solution that capture the key dynamics. Huang et al.~\cite{huang_convective_2023} suggested extracting
\begin{subequations}
    \label{eqn: thermosyphon ode}
    \begin{align}
        X &= \frac{1}{A_0} \int^{2\pi}_{0} \int^{R_1}_{R_2} r^2 u dr d\theta, \\
        Y &= - \frac{1}{A_0} \int_\Omega x T dA, \\
        Z &= - \frac{1}{A_0} \int_\Omega y T dA.
    \end{align}
\end{subequations}
which are the average angular momentum of the flow $X$, flow horizontal center of mass $Y$, and flow vertical center of mass $Z$, respectively, all normalized by the area of the annulus $A_0$. An example trajectory extracted from a flow with $Ra = 1.0\times10^7$ and $Pr = 4.0$ is shown in Figure \ref{fig: Thermosyphon geometry}, which shares a similar attractor structure with the Lorenz system. Using these measured variables, a set of parameterized equations can be derived in the form of
\begin{subequations}
    \begin{align}
        \dot{X} &= -\text{Ra} \text{Pr} Y - \alpha \text{Pr} X, \\
        \dot{Y} &= -kX(Z-z_1) - \beta Y, \\
        \dot{Z} &= kXY - \beta(Z-z_0),
    \end{align}
\end{subequations}
where the constants $\alpha$, $\beta$, $k$, $z_0$, and $z_1$ are all functions of the radii of the thermosyphon \cite{huang_convective_2023}. In the tutorial, we show that SINDyc can be used to learn this parameterized set of equations directly from data, using a cubic polynomial library and STLSQ. The predicted derivatives learned by SINDyc produce an even more accurate model than the "informed" approached, shown in Figure \ref{fig: thermospyphon ode comparison}. Furthermore, we also show how the nonlinear system learned by SINDy can be ported to a symbolic Maths package such as SymPy~\cite{meurer_sympy_2017}, where a nonlinear system analysis can be carried out to accurately identify the critical Rayleigh numbers where bifurcation occurs (see \textbf{Tutorial 3.4}).

The tutorial also demonstrates how the rank-constrained DMD from~\cite{loiseau_data-driven_2020} can be applied to discover the underlying chaotic attractor dynamics. In this approach, we can extract one DMD mode from the azimuthal velocity field and two DMD modes from the temperature field to obtain the low-dimensional system shown in Figure \ref{fig: thermospyphon ode comparison}. Whilst this coordinate transformation with DMD results in a different time-series dataset compared to the "informed" approach shown in \ref{fig: Thermosyphon geometry}, they both share the same chaotic attractor dynamics (see \textbf{Tutorial 3.5}).

Lastly, the tutorial explores how SINDy can be applied to a challenging PDE identification problem, learning Equation \ref{eqn: thermosyphon equations} directly from data. This problem is particularly difficult as the time derivatives cannot be explicitly defined due to the Poisson equation in Equation \ref{eqn: theromsyphon streamfunction}), the system is spatially variant, and the equations are expressed in polar coordinates. Similar to~\cite{supekar2023learning}, we can define a candidate function library with differential operators, such as the Laplacian operator $\nabla^2 (\cdot)$. With this library, we can learn the governing PDE in the form

 \vspace{1pt}
 
\begin{subequations}
    \small
    \begin{align}
        \frac{\partial \omega}{\partial t}
        + \underbrace{\frac{1}{r} \frac{\partial \psi}{\partial \theta} \frac{\partial \omega}{\partial r} - \frac{1}{r} \frac{\partial \psi}{\partial r} \frac{\partial \omega}{\partial \theta}}_{\boldsymbol{u} \cdot \nabla \omega}
        &= \mathrm{Pr} \underbrace{\left( \frac{\partial^2 \omega}{\partial r^2} + \frac{1}{r} \frac{\partial \omega}{\partial r} + \frac{1}{r^2} \frac{\partial^2 \omega}{\partial \theta^2} \right)}_{\nabla^2 \omega}
        + \mathrm{Pr} \, \mathrm{Ra} \left( \frac{\partial T}{\partial r} \cos \theta - \frac{1}{r} \frac{\partial T}{\partial \theta} \sin \theta \right), \\
        \frac{\partial T}{\partial t}
        + \underbrace{\frac{1}{r} \frac{\partial \psi}{\partial \theta} \frac{\partial T}{\partial r} - \frac{1}{r} \frac{\partial \psi}{\partial r} \frac{\partial T}{\partial \theta}}_{\boldsymbol{u} \cdot \nabla T}
        &= \underbrace{\frac{\partial^2 T}{\partial r^2} + \frac{1}{r} \frac{\partial T}{\partial r} + \frac{1}{r^2} \frac{\partial^2 T}{\partial \theta^2}}_{\nabla^2 T}, \\
        \omega &= -\underbrace{\left( \frac{\partial^2 \psi}{\partial r^2} + \frac{1}{r} \frac{\partial \psi}{\partial r} + \frac{1}{r^2} \frac{\partial^2 \psi}{\partial \theta^2} \right)}_{\nabla^2 \psi}.
    \end{align}
\end{subequations}

 \vspace{1pt}
 
We show that using such a library eliminates the collinearity that arises from learning individual partial derivatives, and we show that the true equation using a non-weak form based method can be accurately recovered (see \textbf{Tutorial 3.2}).
\section{Conclusion}

In this tutorial paper and its accompanying \href{https://github.com/paullililili/SINDy4Engineers}{Github repository}, we introduce the SINDy method and show how SINDy can be implemented, extended, and applied to relevant engineering applications. This ranges from learning parametrizable systems with actuation that can be integrated with model-based controllers \cite{brunton_SINDyc_2016, kaiser_SINDy-MPC_2018}, to formulating noise robust extensions such as weak form SINDy~\cite{messenger_WSINDy-ODE_2021, messenger_WSINDy-PDE_2021}. The tutorial also applies SINDy and its extensions to two engineering applications: a UAV system identification and a chaotic heat exchanger modeling problem. We showcase how commonly faced challenges can be mitigated, for example, how noisy data can initially be smoothed, or how sparsity threshold hyperparameters may be selected using statistical stability criteria~\cite{maddu_stability-selection_2022}.

Whilst we have covered several key SINDy extensions, many more exist that tackle other challenges. For example, Kaptanoglu et al.~\cite{kaptanoglu2021promoting} developed the Trapping SINDy optimizer, which promotes the discovery of systems with bounded trajectories. An alternative line of work adopts a Bayesian approach to the regression problem, performing thresholding based on coefficient likelihood rather than magnitude~\cite{niven2020bayesian, hirsh_UQ-SINDy_2022, fung_Bayesian-SINDy_2025}. Separately, Boninsegna et al.~\cite{boninsegna2018sparse} adapted SINDy to discover stochastic differential equations. We recommend readers to check out the \href{https://github.com/dynamicslab/pysindy}{PySINDy repository}~\cite{silva_pysindy_2020, kaptanoglu2021pysindy}, which contains an extensive list of readily available SINDy implementations in Python.

\section*{Acknowledgments}
This work was jointly funded by UK Research \& Innovation's Industrial Doctoral Landscape Awards (reference number UKRI1983) with Airbus UK, and the National Science Foundation AI Institute in Dynamic Systems (grant number 2112085).

\bibliography{src/References}
\bibliographystyle{unsrt}

\newpage
\appendix

\end{document}